\documentclass[sigconf]{acmart}

\usepackage{amsfonts}
\usepackage{algorithmic}
\usepackage{graphicx,color}
\usepackage{textcomp}
\usepackage{xcolor}
\usepackage{hyperref}

\usepackage{amssymb}

\usepackage{subfigure}
\usepackage{dsfont}
\usepackage{multirow}
\usepackage[ruled, norelsize]{algorithm2e}
\AtBeginDocument{%
  }

\copyrightyear{2024} 
\acmYear{2024} 
\setcopyright{acmlicensed}\acmConference[ICBDC 2024]{2024 9th International Conference on Big Data and Computing}{May 24--26, 2024}{Bangkok, Thailand}
\acmBooktitle{2024 9th International Conference on Big Data and Computing (ICBDC 2024), May 24--26, 2024, Bangkok, Thailand}
\acmDOI{10.1145/3695220.3695226}
\acmISBN{979-8-4007-1820-5/24/05}

\begin{document}

%%
%% The "title" command has an optional parameter,
%% allowing the author to define a "short title" to be used in page headers.
%\title{The Name of the Title Is Hope}
\title{Shrinkage Initialization for Smooth Learning of Neural Networks}

\author{Miao Cheng}
\authornote{Corresponding author: miaocheng@acm.org}
%\email{miaocheng@acm.org}
\orcid{0000-0002-6192-7104}
%\author{G.K.M. Tobin}
%\authornotemark[1]
%\email{webmaster@marysville-ohio.com}
\affiliation{%
  \institution{School of Computer Science	\\
Guangxi Normal University}
%  \streetaddress{P.O. Box 1212}
  \city{Guilin}
  \state{Guangxi}
  \country{China}
%  \postcode{43017-6221}
}
\email{miaocheng@acm.org}

\author{Feiyan Zhou}
\affiliation{%
  \institution{School of Computer Science	\\
Guangxi Normal University}
%  \streetaddress{1 Th{\o}rv{\"a}ld Circle}
  \city{Guilin}
  \state{Guangxi}
  \country{China} %}
}
\email{zhfy@mailbox.gxnu.edu.cn}

\author{Hongwei Zou}
\affiliation{%
  \institution{Division of Information Technology	\\
Chongqing Branch of China Merchants Bank}
%  \streetaddress{1 Th{\o}rv{\"a}ld Circle}
  \city{Chongqing}
  \state{}
  \country{China}	%}
}
\email{hongwei_z@cmbchina.com}

\author{Limin Wang}
\affiliation{%
  \institution{Qualcomm (Shanghai) Co. Ltd.}
  \city{Shanghai}
  \country{China}	%}
}
\email{limin@qti.qualcomm.com}

%%
%% By default, the full list of authors will be used in the page
%% headers. Often, this list is too long, and will overlap
%% other information printed in the page headers. This command allows
%% the author to define a more concise list
%% of authors' names for this purpose.
\renewcommand{\shortauthors}{Cheng et al.}

%%
%% The abstract is a short summary of the work to be presented in the
%% article.
\begin{abstract}
The successes of intelligent systems have quite relied on the artificial learning of information, which lead to the broad applications of neural learning solutions. As a common sense, the training of neural networks can be largely improved by specifically defined initialization, neuron layers as well as the activation functions. Though there are sequential layer based initialization available, the generalized solution to initial stages is still desired. In this work, an improved approach to initialization of neural learning is presented, which adopts the shrinkage approach to initialize the transformation of each layer of networks. It can be universally adapted for the structures of any networks with random layers, while stable performance can be attained. Furthermore, the smooth learning of networks is adopted in this work, due to the diverse influence on neural learning. Experimental results on several artificial data sets demonstrate that, the proposed method is able to present robust results with the shrinkage initialization, and competent for smooth learning of neural networks.
\end{abstract}

\ccsdesc[500]{Computing methodologies~Neural networks}
\ccsdesc[500]{Theory of computation~Dynamic programming}

%%
%% Keywords. The author(s) should pick words that accurately describe
%% the work being presented. Separate the keywords with commas.
\keywords{Neural networks, Initialization, Shrinkage approach, Smooth learning}
%% A "teaser" image appears between the author and affiliation
%% information and the body of the document, and typically spans the
%% page.
%\begin{teaserfigure}
%  \includegraphics[width=\textwidth]{sampleteaser}
%  \caption{Seattle Mariners at Spring Training, 2010.}
%  \Description{Enjoying the baseball game from the third-base
%  seats. Ichiro Suzuki preparing to bat.}
%  \label{fig:teaser}
%\end{teaserfigure}

%\received{20 February 2007}
%\received[revised]{12 March 2009}
%\received[accepted]{5 June 2009}

%%
%% This command processes the author and affiliation and title
%% information and builds the first part of the formatted document.
\maketitle

\section{Introduction}
The emerge of digital life has triggered a new era of intelligent and ubiquitous computing \cite{Zhou14BigData}\cite{Bengio13RL}, leading to success of neural learning solutions, e.g., neural networks \cite{Hinton06NN}\cite{Hastie11ESL}, deep learning \cite{Bishop23DLFC}\cite{Goodfellow2016DL}. With the multilayer structures of perceptron, the well-known neural networks are able to learn the optimal networks for forward inference of information, and the fine data can be obtained in accordance with the targets that are to be approximated \cite{Bishop23DLFC}\cite{Glorot10DFNN}\cite{Haykin16NNLM}. Derived from neural learning, deep learning is able to achieve the similar function as the networks, and the learning ability can be promised to be reached with the optimized learning of deep refinements \cite{Aggarwal23NNDL}. Nevertheless, the huge calculation complexity has prohibited it from efficiency, owing to the training of complicated structures of networks \cite{Taylor17MNN}\cite{Taylor17NNVIB}\cite{Cuda00PC}. As a consequence, a long term of duration is necessary for a series of forward and backpropagation stages, and it seems that such dilemma cannot be avoided as a common. Without loss of generality, the networks are organized as several layers of learning perceptron, where each layer consists of the connections of multiple neurons with the next layer \cite{Hastie11ESL}\cite{Chan15PCANet}. Obviously, the goal of neural learning is to optimize the connections of each layer of networks to approach the targets of each input \cite{Krizhevsky12CNN}\cite{Sercu16DCNN}. Normally, the optimal outputs can be obtained if enough training epochs are paid, and the best approximation is to be reached. In addition, it can be resorted to be temporal extension of stream learning with recurrent connections \cite{Bengio94RNN}.
% and temporal memory \cite{Hochreiter97LSTM}. 

In terms of the exhausted complexity of training of networks which is the intrinsic limitation of neural learning, there are two main categories of solutions that have been proposed and adopted broadly \cite{Bishop11PRML}. The challenge of the first category of the state-of-the-art solutions have been conducted as the relaxation of outputs of each layer, benefiting from the sparseness of resulting valid neurons \cite{Hinton15DKNN}\cite{Bengio07GT}. As a consequence, the dropout stage is appended to each neural layer of networks, and certain ratios of neurons are randomly selected to be null to accelerate the learning speed of training \cite{Srivastava14Dropout}. In addition, the dropout stage can also ensure convergence of optimization of neural learning according to the practical outputs, while light complexity is required for universal computing. The second category of optional methods are to optimize the initialization of networks, and ideal learning of neurons can be expected with the good start of training approach \cite{Ioffe15BN}\cite{Saxe14DIN}\cite{Mishkin16LSUV}. Distinguishingly, the initialization of networks is unnecessary to be performed in each training of epochs, and single one piece of optimization is desired in the beginning. Furthermore, the calculation complexity can be controlled with respect to the one circle of learning, while the improved networks can be inferred for the optimized learning of training approach. Though the improvements are limited with diverse categories of data, the training stage can be optimized and changed to be better for inference of networks \cite{Graham14FMP}\cite{Murray14GMP}. As important steps of neural networks, activation functions aim to transform the inputs from the previous layer into the normalized outputs, which is another important issue of networks that promise it to be stable for optimization \cite{Bishop23DLFC}\cite{Goodfellow2016DL}. The most popular activation functions can be referred to Sigmoid, Tanh and ReLU functions, which are defined as
%\begin{equation}
%  f\left( x \right) = {\rm Sigmoid}  \left( x \right) = \frac{1}{{1 + {e^{ - x}}}}
%\end{equation}
%\begin{equation}
%  f\left( x \right) = {\rm tanh}  \left( x \right) = \frac{{{e^x} - {e^{ - x}}}}{{{e^x} + {e^{ - x}}}}
%\end{equation}
%\begin{equation}
%  f\left( x \right) = {\rm ReLu} \left( x \right) = \max \left( {0, x} \right)
%\end{equation}
\begin{equation}
  {\rm Sigmoid}  \left( x \right) = \frac{1}{{1 + {e^{ - x}}}}
\end{equation}
\begin{equation}
  {\rm Tanh}  \left( x \right) = \frac{{{e^x} - {e^{ - x}}}}{{{e^x} + {e^{ - x}}}}
\end{equation}
\begin{equation}
  {\rm ReLU} \left( x \right) = \max \left( {x, 0} \right)
\end{equation}
Other activation functions are also proposed to enhance the forward stages of networks or accelerate the learning speed of the derivatives of each layer \cite{Goodfellow13Maxout}\cite{Clevert16ELU}. Note that, the influence of activation functions can be quite large for the optimization of each layer of networks, due to the different distributions of outputs of each layer, especially for the nonsmooth ones \cite{Chang15BN}. Particularly, it can be the keypoint of learning ability, and that is the reason why the smooth activation function is selected in this work to reach a correct evaluation of different methods.

As a successful solution to initialization of networks, batch normalization (BN) aims to transform the outputs of each layer to be with the normalized variances, which are pushed as the inputs of the neurons of the next layer \cite{Glorot10DFNN}\cite{Ioffe15BN}. Due to the simple implementation of normalization, BN is able to afford the ubiquitous computing of neural learning, and further advances are extensible for incremental optimization \cite{Cheng19SPA}\cite{Cheng22TANMF}\cite{Cheng10LDSE}. Another attempts have conducted the initialization of network as a standard function of connections of different layers, and consequently, the optimized transformations are referred to seek for the ideal setting of inputs. Thereafter, the optional choice of initialization mainly depends on the understanding of connections of neurons of layers, and reasonable transformations can be expected for the inference of neural learning \cite{Glorot10DFNN}\cite{Chang15BN}. In terms of the idea, the most outstanding method has been devised to optimize the bridge between the previous and the next layers, and brightness of inspiration can be attained for the correspondence of smooth touches \cite{Glorot10DFNN}\cite{Liao16NL}. Nevertheless, the original solution has been prevented from the simple perceptron that consists of a few layers, and particularly, the specific conditions have been assumed to be promised in fact. As a consequence, it is hardly to be extended to multiple layers of networks, and generalized initialization is still desired for the discovery of common structures of networks. Furthermore, the complexity of initialization is still necessary to be controlled strictly, and the batch approach is to be adopted as a promise.

In terms of those issues, an improved initialization approach to smooth learning of networks is presented in this work, while shrinkage initialization is devised based on the bridges of neurons. Distinguishing from the existing methods, the proposed method holds a generalized framework and is able to initialize the networks with any quantity of layers. Furthermore, the normalized skeleton of median layer is pushed to enhance the invariant transformation of networks. The rest of this paper is organized as follows. Some backgrounds of the related works are given in section 2. The main idea of the proposed method is given in section 3. The experimental results are disclosed in section 4. Finally, the conclusion is draw in section 5.

\section{Background}
Though the structures of networks can be quite complicated, the conception of a neural network is straightforward for depiction of learning stages. More specifically, the input data $ X \in \mathds{R}^{d \times n} $ are pushed into the first layer of the network, and a linear transformation with certain  weights is assigned while the obtained results are transferred to the next layer as the input data \cite{Goodfellow2016DL}\cite{Haykin16NNLM}, e.g.,
\begin{equation}
  g\left( x \right) = wx + b
\end{equation}
Here, $ w $ indicates the transformation that transforms $ x $ into the target representation, $ b $ denotes the bias parameter. 
After that, the resulting data are normally filtered by an activation function, which holds the power of smoothness of data while normalization can be attained. Then, such approach is to be repeated once again and again, till the final layer of network is reached. Furthermore, it has been disclosed that, though suitable fits of network can be ideal, more deeper layers do not always lead to better performance. Nevertheless, it has been a common sense any layers of network are necessary to be optimized with a backpropagation approach. Such approach traverses the network from the final layer to the front one and optimizes the weights of each layer with certain update step that are learned based on the previous layer of the backward direction.

The true fact about initialization of networks is that the random initialization of each layer is able to reach good results if enough epochs are given for neural learning. Nevertheless, an ideal initialization is to promise to present acceleration of optimization speed of networks, which is approach to the matching of network structures with respect to fixed data. As a consequence, the difficulty of initialization of networks has been explained as the overfitting of networks and the loss of smoothness of input data for the next layer. With the empirical observations, the models of network assume a balanced initial distribution of data with respect to the domain of the piecewise linear activation function \cite{Liao16NL}. Thus, the batch normalization is brought into deep feedforward neural networks, where each region of the activation function is trained with a relatively large proportion of training samples. As the most popular initialization of networks, batch normalization (BN) can be simply grafted into the layers of any networks. The basic idea of BN is to normalize each layer of data into unit variance statistically, and smooth results can be achieved accordingly. The limitation of this approach is that the normalized data of each layer are still independent from forward and backpropagation steps of networks, while the subnetworks share their parameters with other subnetworks definitely.

% &&&&&&&&&&&&&&&&&&&&&&&&&&&&&&&&&&&&&&&&&&&&&&&&&&&&&&&
\begin{figure*}
    \centering
    %\subfigure[]
	{ \includegraphics[width=0.8\textwidth]{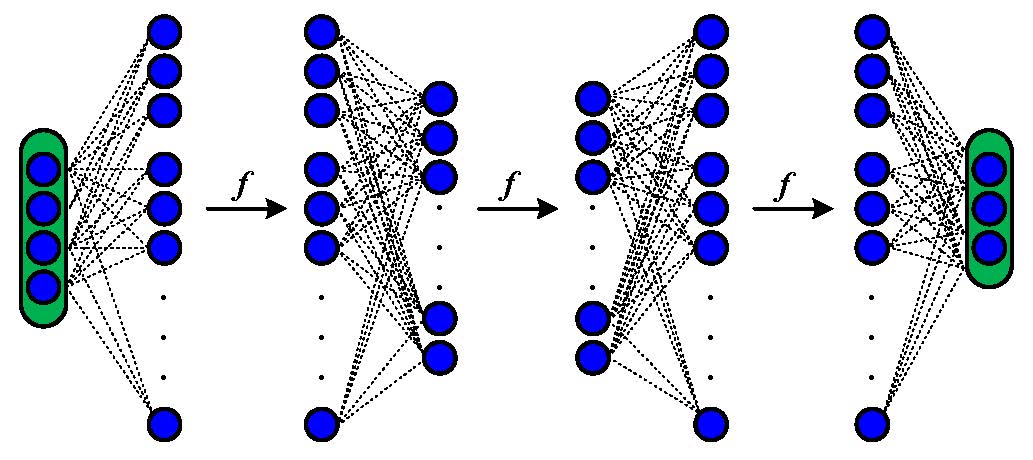}	}
    \caption{The illustration of neural learning of network. }
\end{figure*}
% &&&&&&&&&&&&&&&&&&&&&&&&&&&&&&&&&&&&&&&&&&&&&&&&&&&&&&&
In terms of this, dynamic initialization of neural learning (DIN) \cite{Saxe14DIN} and the layer-sequential unit-variance (LSUV) \cite{Mishkin16LSUV} initialization are proposed for weight initialization of deep nets. Firstly, it initializes the weights of each convolution or inner product layer with orthonormal transformation. Secondly, the normalization is proceeded from the first to the final layer by normalizing the variance of the output of each layer. Rather than combinatorial learning \cite{Kuo22DSDNet}, the orthogonalization of each layer is restricted to the single layer of itself, while correspondence of layers is still ignored. Furthermore, the orthogonalization is extended to the connection of pairs of layers of networks by updating the weights of each pair of nets simultaneously. More specifically, the front and the back layers are to be updated with the multiply of orthogonal matrices derived from the singular value decomposition (SVD) of data \cite{Golub13MC}, e.g.,
\begin{equation}
 {{\rm E}_{31}} = {U_{33}}{S_{31}}V_{11}^T
\end{equation}
where $ {\rm E}_{31} $  is the correlation matrix associated with the third and the first layers that is calculated as
\begin{equation}
 {{\rm E}_{31}} = {X_3}X_1^T
\end{equation}
Here, $ X_3 $ indicates the aligned data of the third layer, while $ X_1 $ indicates the input data from the front layer. As a consequence, the linear transformation of networks can be updated as
\begin{equation}
 {W_{21}} = {\overline W _{21}}V_{21}^T
\end{equation}
\begin{equation}
 {W_{32}} = {U_{33}}{\overline W _{32}}
\end{equation}
The pain of such intuition normally suffers from the fixed linear transformation of networks, which leads to the theoretical validation of correspondence of network units. Furthermore, the update of weights is prevented from extension of broad nets with the complicated structures, while orthogonal rotation is adopted definitely. Another issue about forward learning of networks is the dropout stage, which sets certain ratio of elements of outputs to be null. Since it is natural to optimized learning of neural units, and demonstrated to be stable intuitively, it is adopted in this work straightforward.

\section{Shrinkage Initialization of Neural Learning}
The fully connected networks, also known as multi-layer perception machine, are competent to learn the matched neural structures with input data, while all neural units are connected with the previous and back layers. Benefiting from the absorption of common parameters of units, the traits of characteristics can be communicated between the connected units. Without loss of generally, there are still some issues that should be highlighted. Firstly, the activation functions can be optional for each layer, resulting in different speed of convergency and outputs. That is, the smooth activation is intuitive for inference of linear units, while nonsmooth activation can also be fit with targets specifically. As a consequence, it is hardly to conclude whether the learned results are benefited from either side of incoming. Furthermore, the generalized structures of networks are more common with random layers of networks, and thus, the extensions of initialization can be derived accordingly.

%% &&&&&&&&&&&&&&&&&&&&&&&&&&&&&&&&&&&&&&&&&&&&&&&&&
%\begin{figure*}
%\centering
%\subfigure[]{
%\includegraphics[width=0.19\textwidth]{./Figure/Fig2/non_w2_1}}
%\subfigure[]{
%\includegraphics[width=0.19\textwidth]{./Figure/Fig2/bn_w2_1}}
%\subfigure[]{
%\includegraphics[width=0.19\textwidth]{./Figure/Fig2/din_w2_1}}
%\subfigure[]{
%\includegraphics[width=0.19\textwidth]{./Figure/Fig2/lsuv_w2_1}}
%\subfigure[]{
%\includegraphics[width=0.19\textwidth]{./Figure/Fig2/sinl_w2_1}}
%\caption{The learned transformation of the last layer of network from different initialization methods on Coil 20 data set. (a) None (b) BN (c) DIN (d) LSUV (e) SINL. }
%\end{figure*}
%% &&&&&&&&&&&&&&&&&&&&&&&&&&&&&&&&&&&&&&&&&&&&&&&&&
% &&&&&&&&&&&&&&&&&&&&&&&&&&&&&&&&&&&&&&&&&&&&&&&&&
\begin{figure*}
\centering
\subfigure[]{
\includegraphics[width=0.24\textwidth]{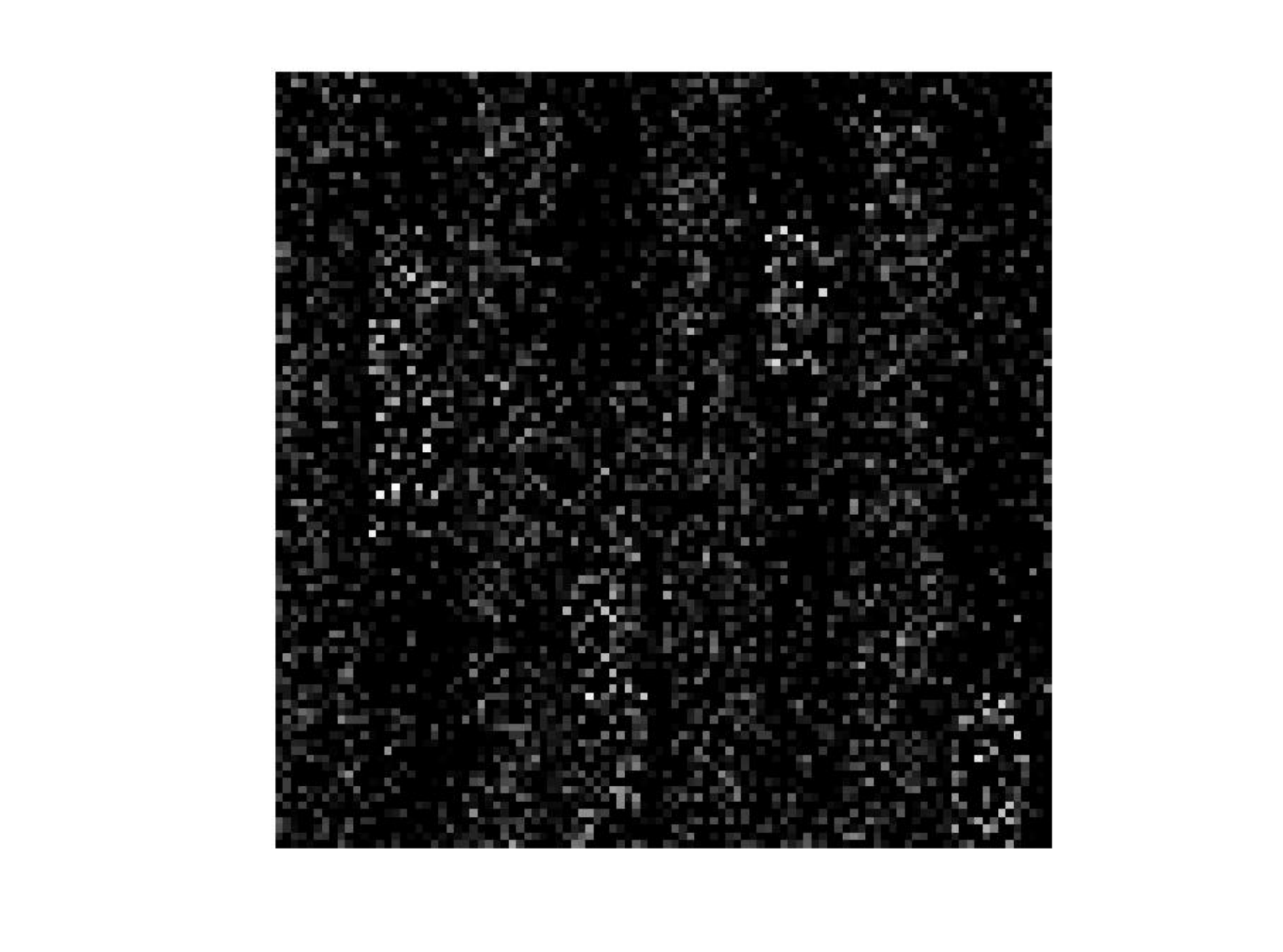}}
%\subfigure[]{
%\includegraphics[width=0.19\textwidth]{./Figure/Fig2/sinl_w2_4k}}
\subfigure[]{
\includegraphics[width=0.24\textwidth]{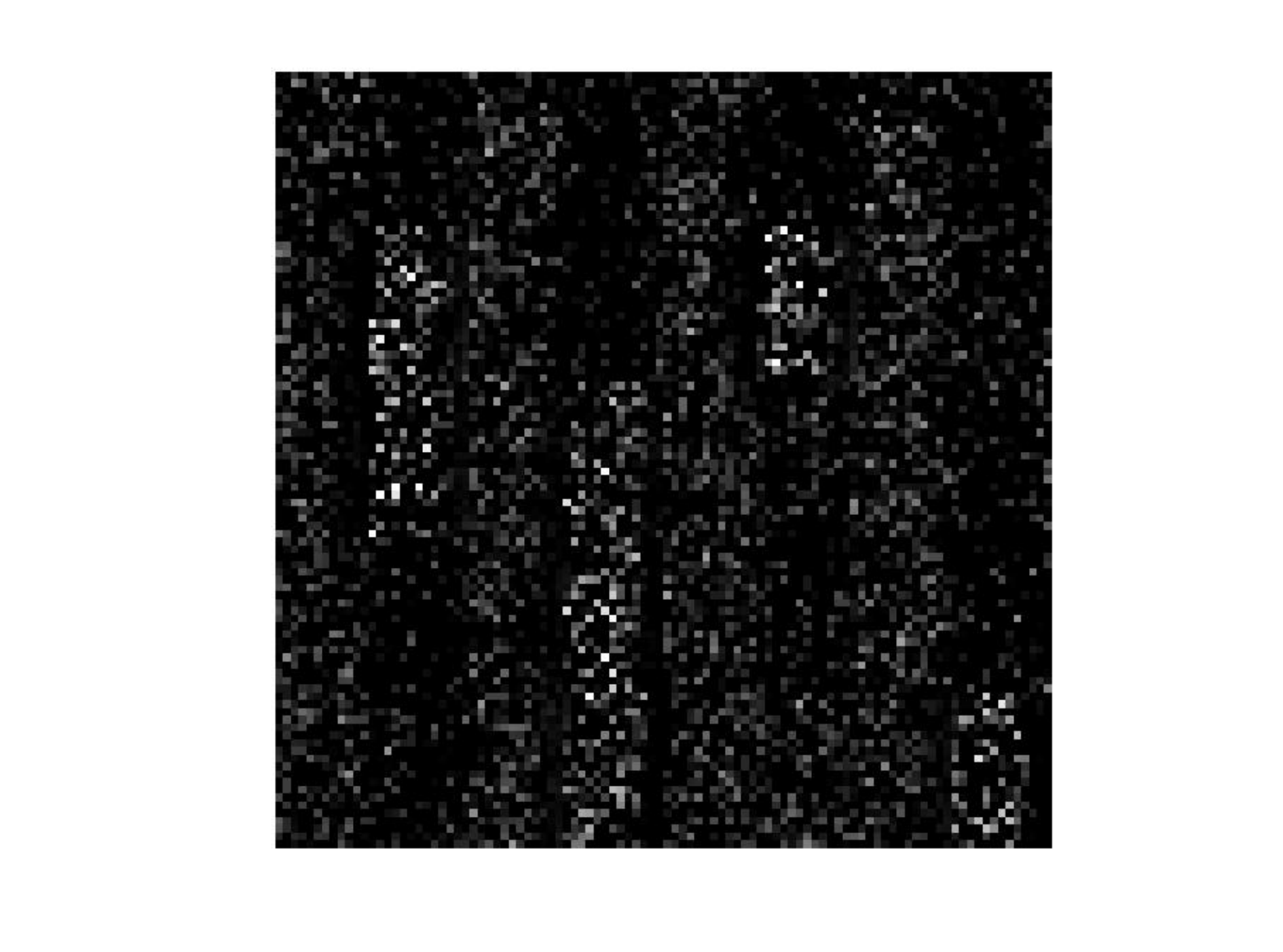}}
\subfigure[]{
\includegraphics[width=0.24\textwidth]{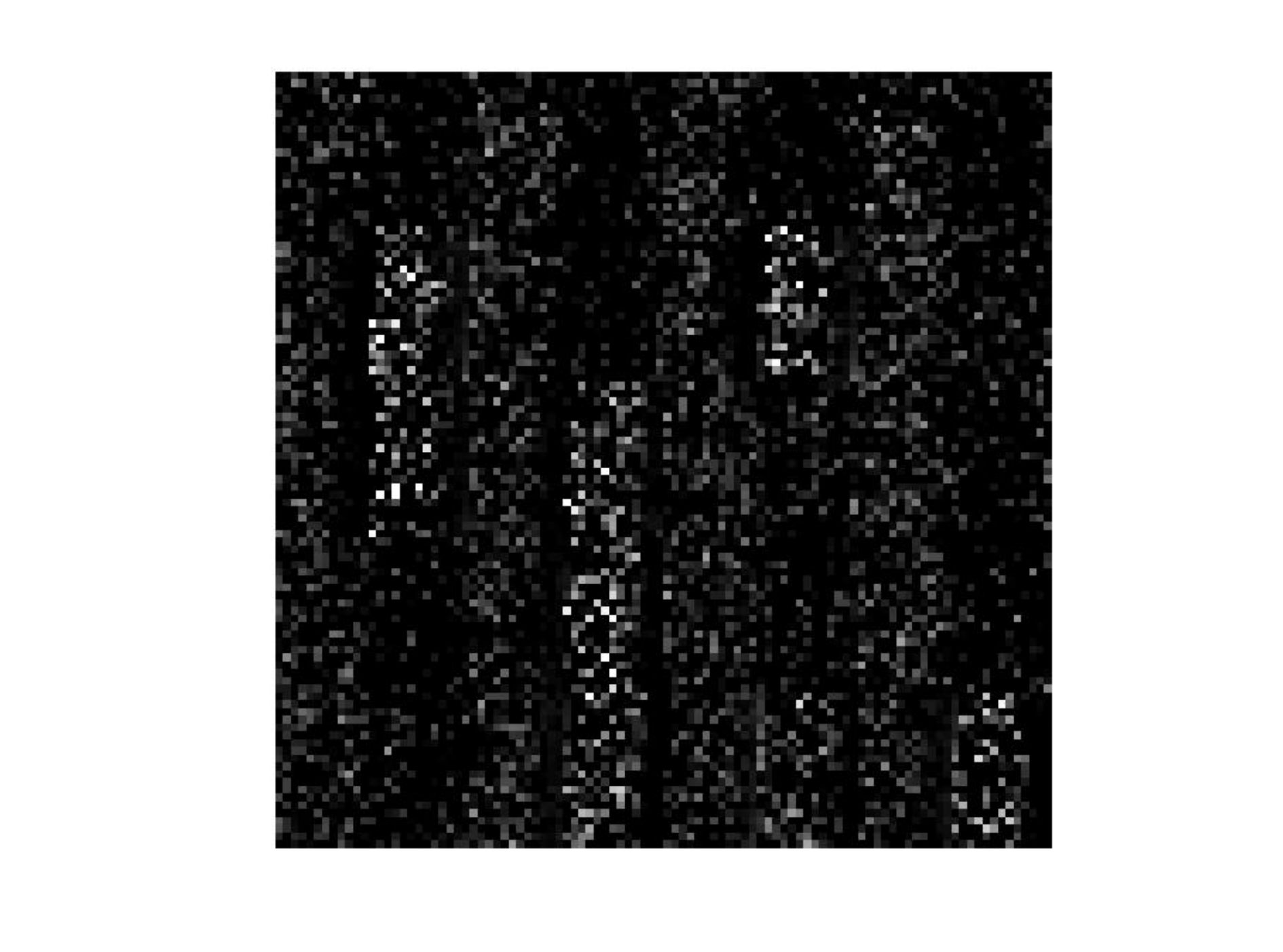}}
\subfigure[]{
\includegraphics[width=0.24\textwidth]{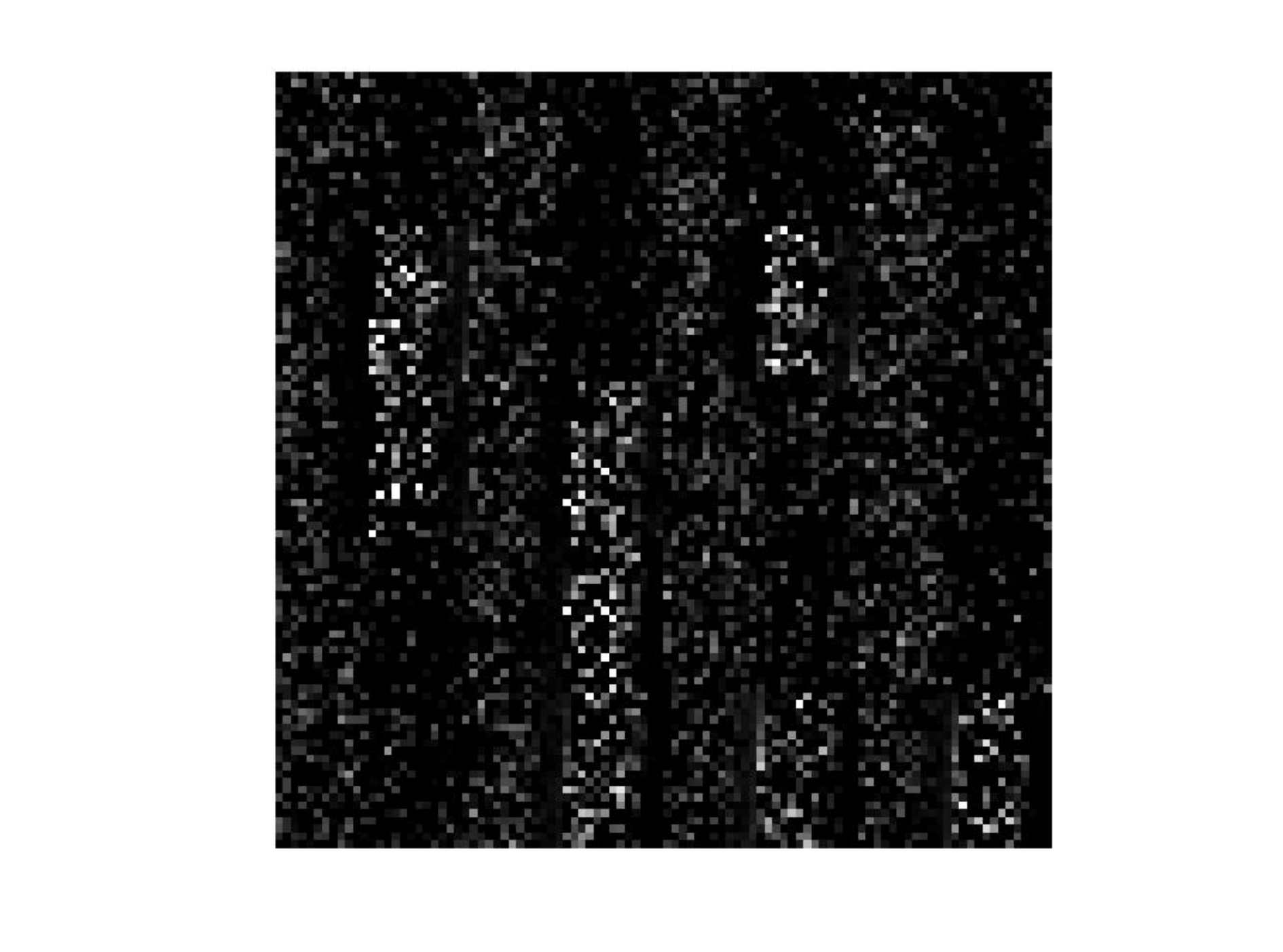}}
\caption{The learned transformation of network derived from shrinkage initialization based on different epoches. (a) 2000 (b) 4000 (c) 6000 (d) 8000. }
\end{figure*}
% &&&&&&&&&&&&&&&&&&&&&&&&&&&&&&&&&&&&&&&&&&&&&&&&&
Assume that the transformation of each layer is given randomly, while the input data of each layer can be inferred beforehand. Thereafter, the connected bridge between either side can be defined as the linear transformation, e.g.,
\begin{equation}
  {X_j} = {W_{i \to j}}{X_i}
\end{equation}
Conceptually, the $ {W_{i \to j}} \in {\mathds{R}^{q \times p}} $ denotes the total linear transformation that transforms $ X_i $ to $ X_j $. Note that, if the single transformation is referred, the ideal $ W_{i \to j} $ can be actually calculated as 
\begin{equation}
 {\rm E}_{ij} = {X_j}X_i^T{\left( {{X_i}X_i^T} \right) }^ {+ }
\end{equation}
Then, the orthogonal matrices of the affinity of layers can be calculated with the SVD of $ {\rm E}_{ij} $, e.g.,
\begin{equation}
  {{\rm E}_{ij}} = {U_{ij}}{S_{ij}}V_{ij}^T
\end{equation}
As a consequence, the transformation of weights of the specific neural units can be updated as
\begin{equation}
  {W_{i \to i + 1}} = {W_{i \to i + 1}}V_{ij}^T
\end{equation}
\begin{equation}
 {W_{j - 1 \to j}} = {U_{ij}}{W_{j - 1 \to j}}
\end{equation}
Note that, both $ {U_{ij}} \in {\mathds{R}^{q \times q}} $ and $ {V_{ij}} \in {\mathds{R}^{p \times p}} $ are full-rank orthogonal matrices with the square shape. Nevertheless, it is able to be calculated efficiently due to the small shape of transformation \cite{Cheng11MMC}\cite{Cheng18MMC}. Furthermore, it is noticeable that, the orthogonal matrices actually push a rotation of the original transformation, while the characteristics of transferring can be reserved. That is, it is not the exact matching results for correspondence, but adjustments are competent for initialization of neural networks. Besides, the update is performed from the boundary sides of networks, and the median layer is approached stepwise, which can be adaptable for generalized structures of networks.

% &&&&&&&&&&&&&&&&&&&&&&&&&&&&&&&&&&&&&&&&&&&&&&&&&&&&&&
\begin{algorithm}
\caption{Shrinkage Initialization of Neural Learning}
\KwIn{ The input data $ X \in \mathds{R}^{d \times n} $, the quantity of layers $ m $, the dimensionality of each layer, the defined activation function, as well as the parameters that are adopted in networks. }
\KwOut{The initialized network.}
1. Randomly initialize the transformation of each layer.		\\
2. Calculate the resulting data of each layer based on the forward approach of network.	\\
3. \While{ The median layer has never been reached }
	{
		4. Calculate the independent bridge between the current data of the boundary layers, and obtain $ {\rm E}_{ij} $.		\\
		5. Calculate SVD of $ {\rm E}_{ij} $, and obtain the orthogonal matrices $ U_{ij} $ and $ V_{ij} $ respectively.	\\
		6. Update the transformation $ W_{i \to { i + 1 } } $ and $ W_{ {j - 1} \to j} $ of the boundary layers with orthogonal rotations, respectively.	
	}
7. \If { The quantity of layers is odd }
	{
		8. Calculate SVD of the transformation   of the median layer, and update it with the normalized reconstruction. 
	}
9. Perform the batch normalization if required.
\end{algorithm}
% &&&&&&&&&&&&&&&&&&&&&&&&&&&&&&&&&&&&&&&&&&&&&&&&&&&&&&
Nevertheless, it is noticeable that, the median layer is to be suspended for initialization, due to the fact that, the quantity of neural layers is randomly set that may lead to the odd number. In terms of such issue, the independent decomposition of the linear transformation of median layer is adopted, e.g.,
\begin{equation}
  {W_{i \to j}} = {U_{ij}}{S_{ij}}V_{ij}^T
\end{equation}
Thereafter, it is to be simply updated as the reconstruction of normalized orthogonal matrices, e.g.,
\begin{equation}
 {W_{i \to j}} = {U_{ij}}V_{ij}^T
\end{equation}
Instead of the original transformation, the unit orthogonal matrix can be competent for the normalized transformation, while the main characteristics of matching can be reserved. Obviously, the main idea of the proposed initialization is to improve the weights with the orthogonal rotations, while the correspondence between each pair of network units can be reserved. The obtained observations can be summarized as several issues. Firstly, the orthogonal update mechanism is adopted to the networks that consist of a few layers, and the extension of generalized networks are still desired. Furthermore, the orthogonal rotation is to be corresponding to the different pairs of layers, while the elastic matching is necessary to be absorbed into optimization. Furthermore, the batch normalization is also available for the proposed SINL approach as an attachment for initialization. Without loss of generality, the whole procedure of the proposed initialization approach can be summarized stagewise, which is given in algorithm 1.

In addition, there are some analyses that are necessary to be highlighted. Obviously, the complexity of the proposed shrinkage initialization is mainly based on the quantity of layers. And the complexity of each iteration mainly depends on the decomposition of the bridge $ {W_{ij}} \in {\mathds{R}^{q \times p}} $, such as $ O \left( {{p^2}q + p{q^2}} \right) $, while the total complexity of initialization depends on the half quantity of layers. Furthermore, the cost of the bridge between $ {X_i} \in {\mathds{R}^{p \times n}} $ and $ {X_j} \in {\mathds{R}^{q \times n}} $ requires $ O \left( {pqn + {q^3}} \right) $ for the inverse and multiply calculation of data. In summary, the complexity of initialization mainly depends on the quantity of the layers of neural networks and the shape of transformation aligned with each layer. Note that, the shape of each layer is still small commonly, and can be inferred efficiently.

% &&&&&&&&&&&&&&&&&&&&&&&&&&&&&&&&&&&&&&&&&&&&&&&&&
\begin{figure*}
\centering
\subfigure[]{
\includegraphics[width=0.32\textwidth]{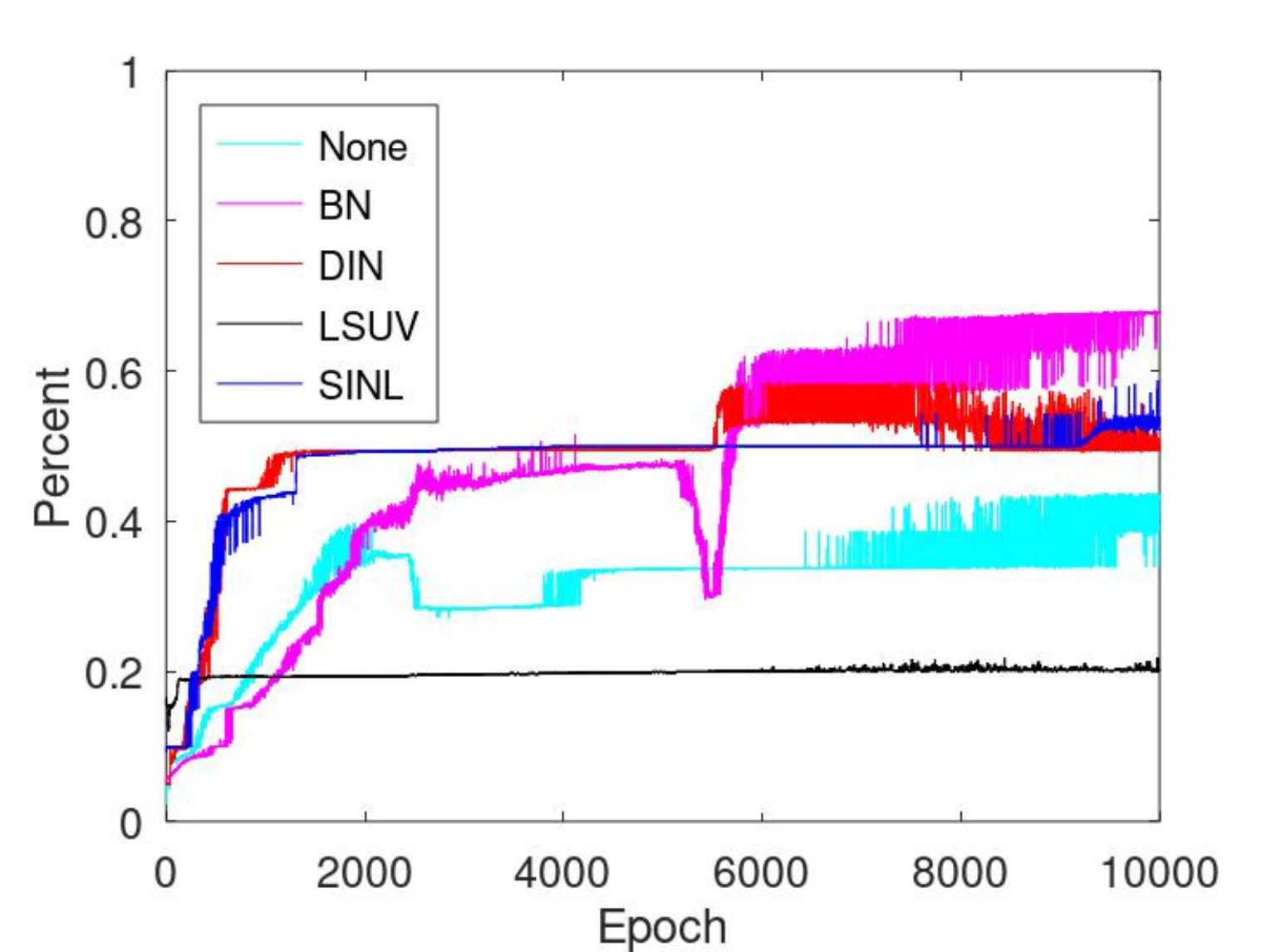}}
\subfigure[]{
\includegraphics[width=0.32\textwidth]{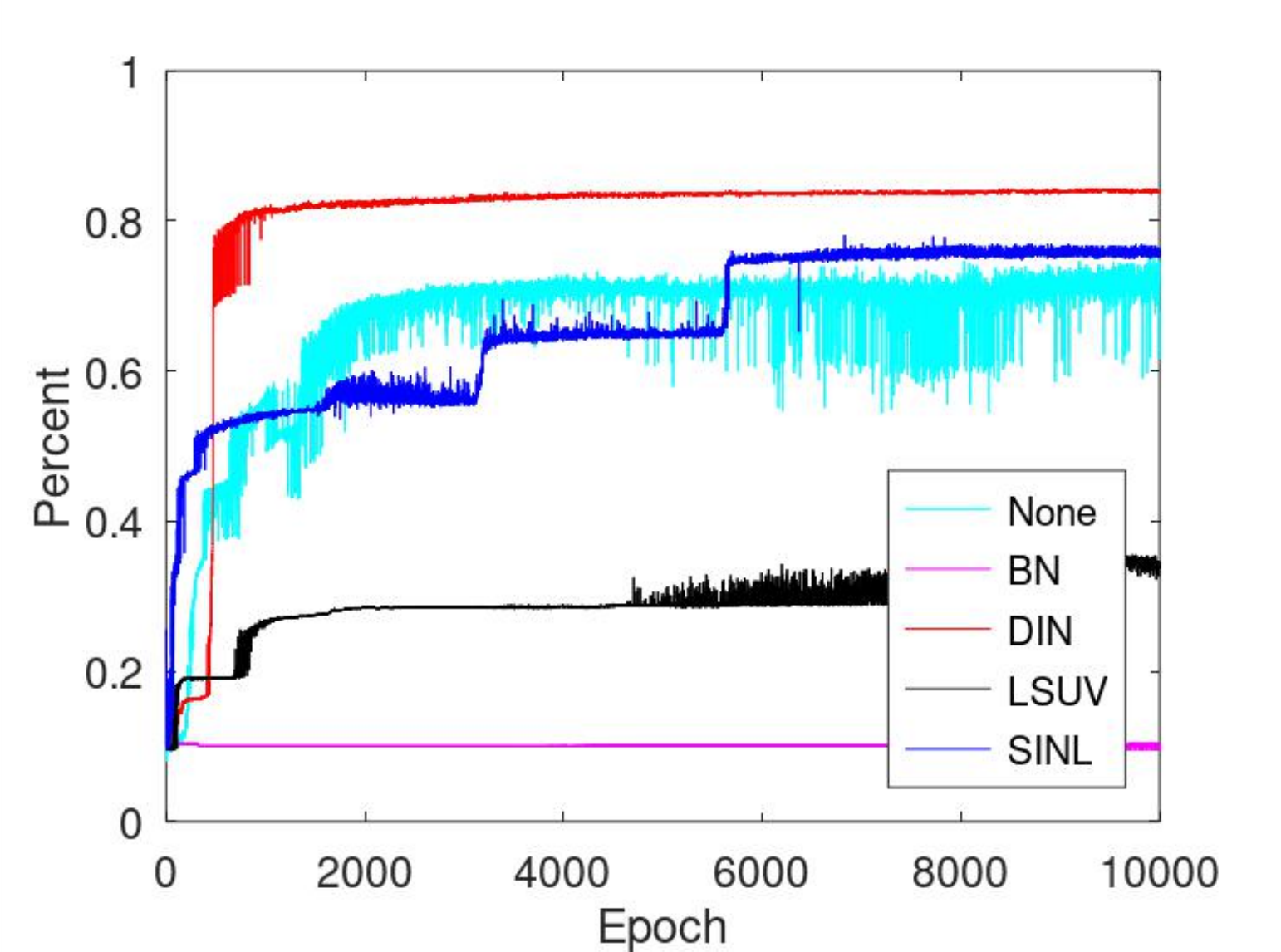}}
\subfigure[]{
\includegraphics[width=0.32\textwidth]{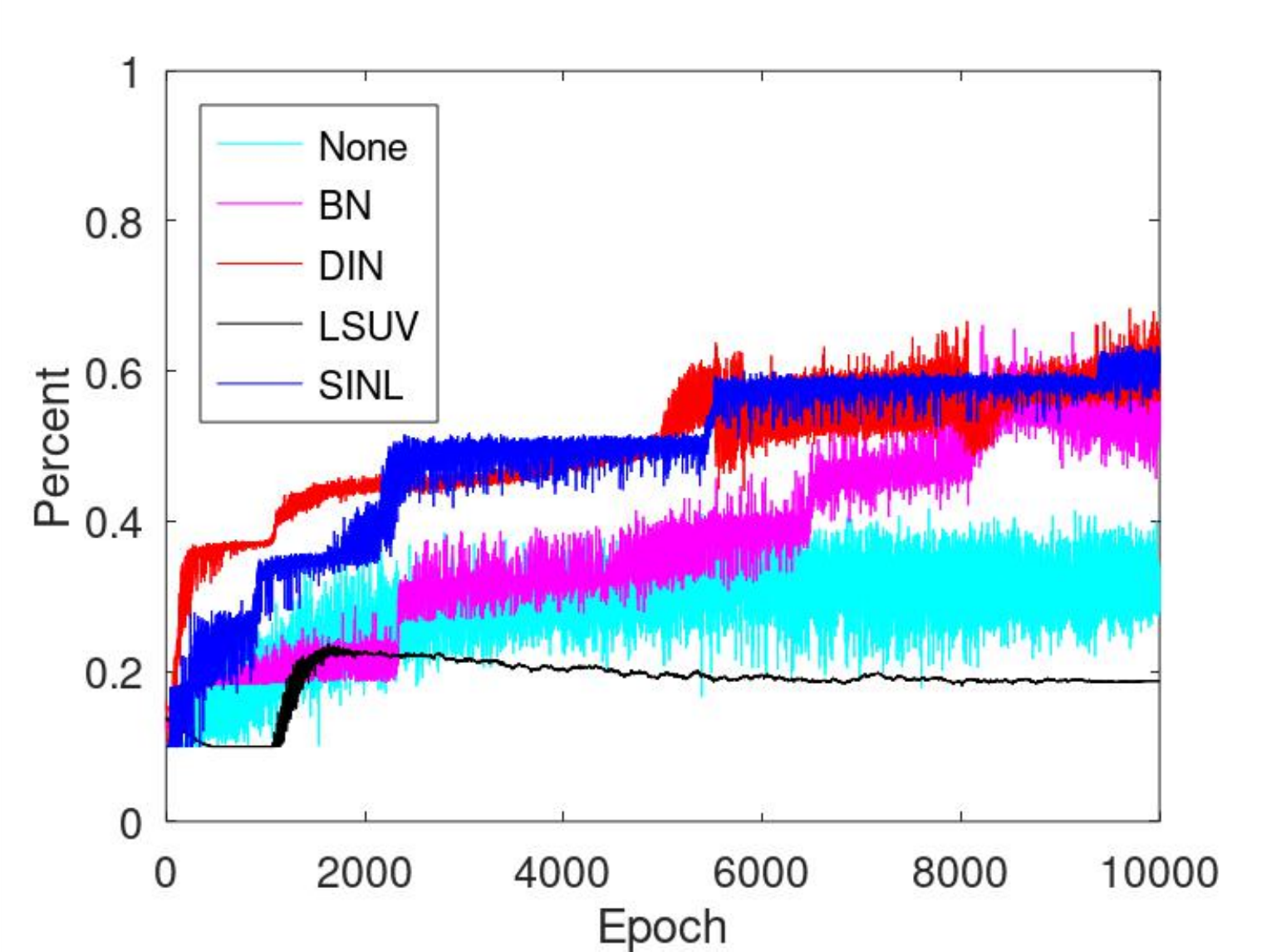}}
\caption{ The obtained accuracy associated with the iterative epochs on the different data sets. (a) Coil 20 (b) Monkey (c) Letter. }
\end{figure*}
% &&&&&&&&&&&&&&&&&&&&&&&&&&&&&&&&&&&&&&&&&&&&&&&&&
% &&&&&&&&&&&&&&&&&&&&&&&&&&&&&&&&&&&&&&&&&&&&&&&&&
\begin{figure*}
\centering
\subfigure[]{
\includegraphics[width=0.32\textwidth]{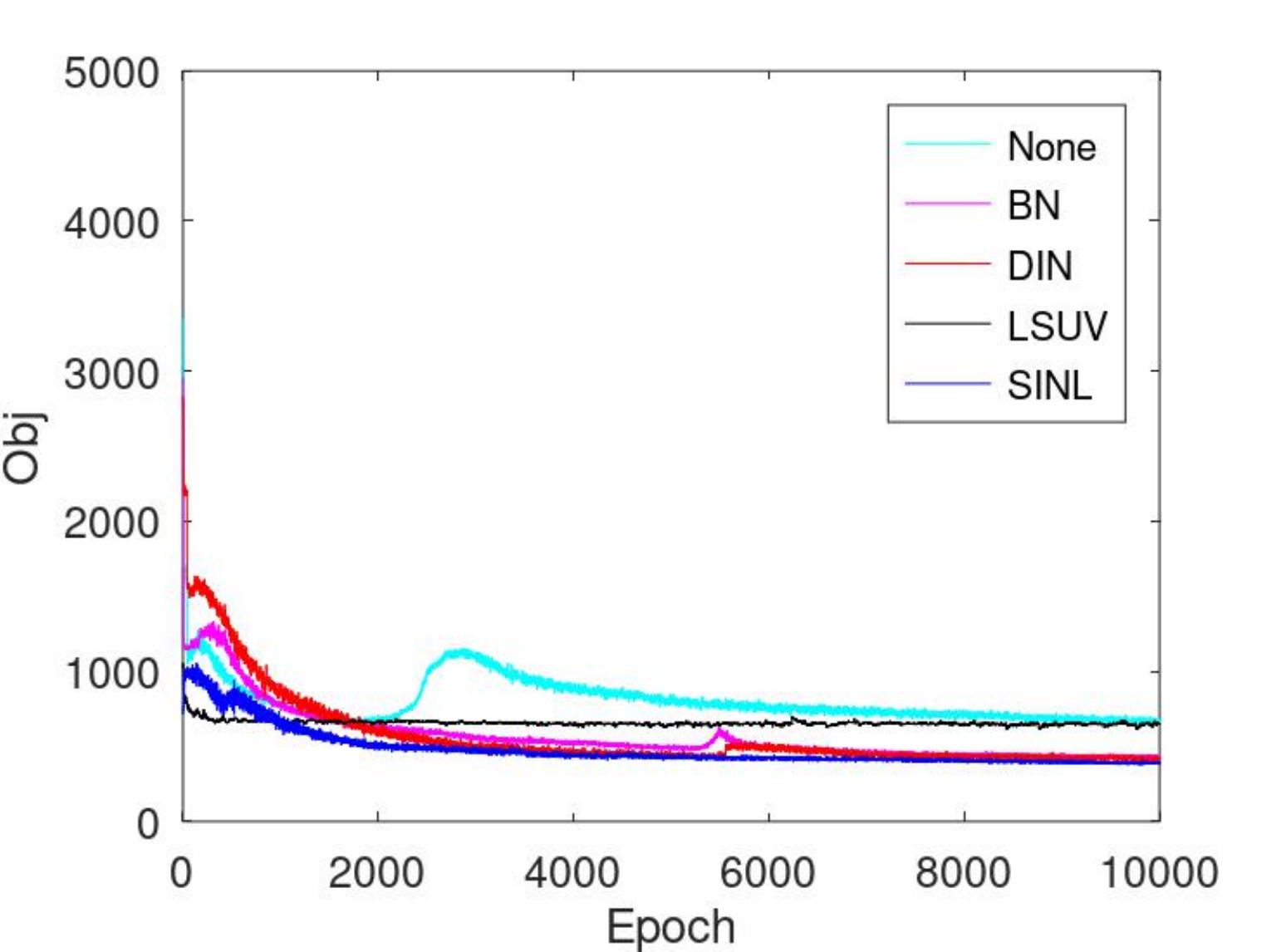}}
\subfigure[]{
\includegraphics[width=0.32\textwidth]{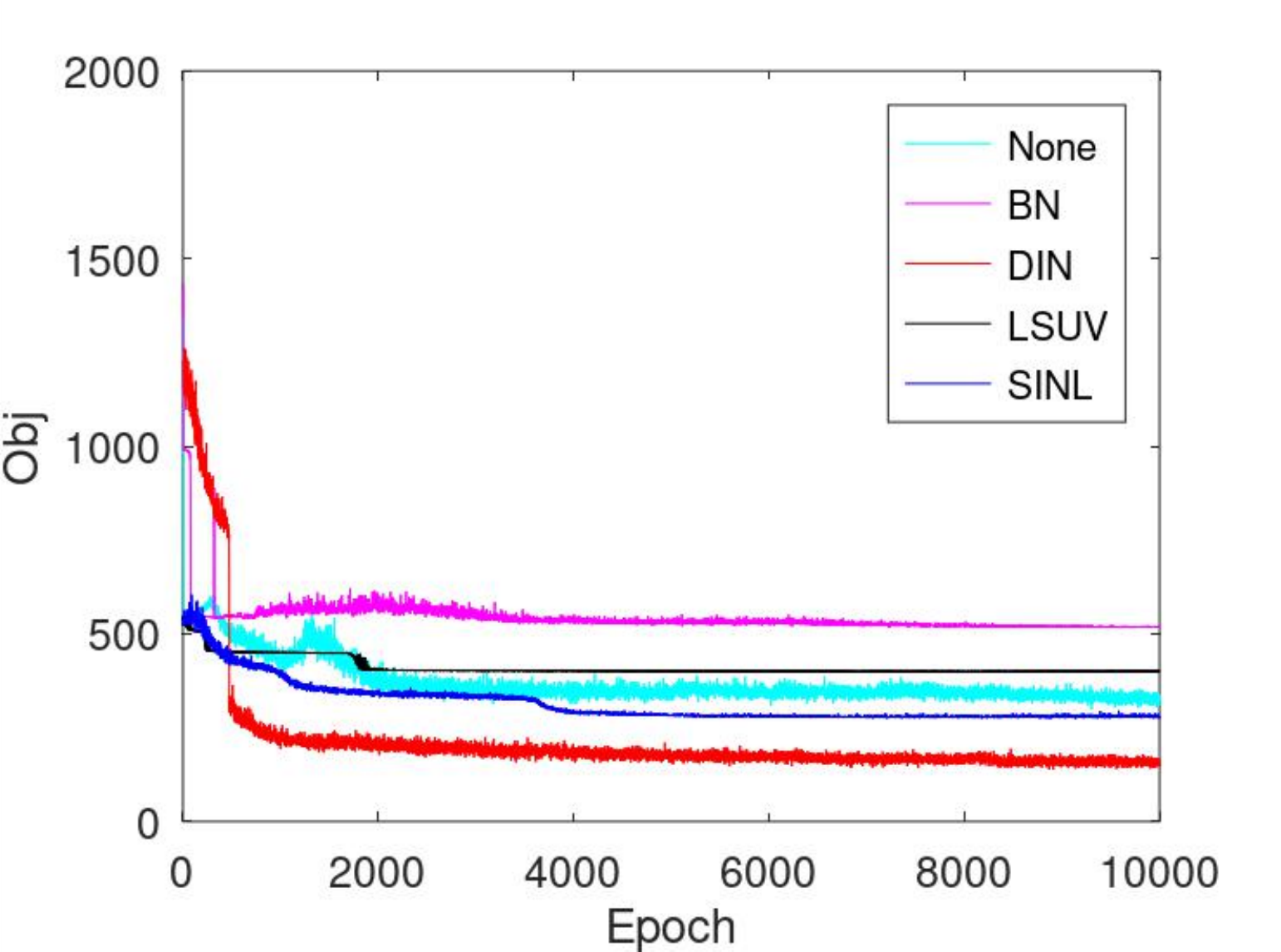}}
\subfigure[]{
\includegraphics[width=0.32\textwidth]{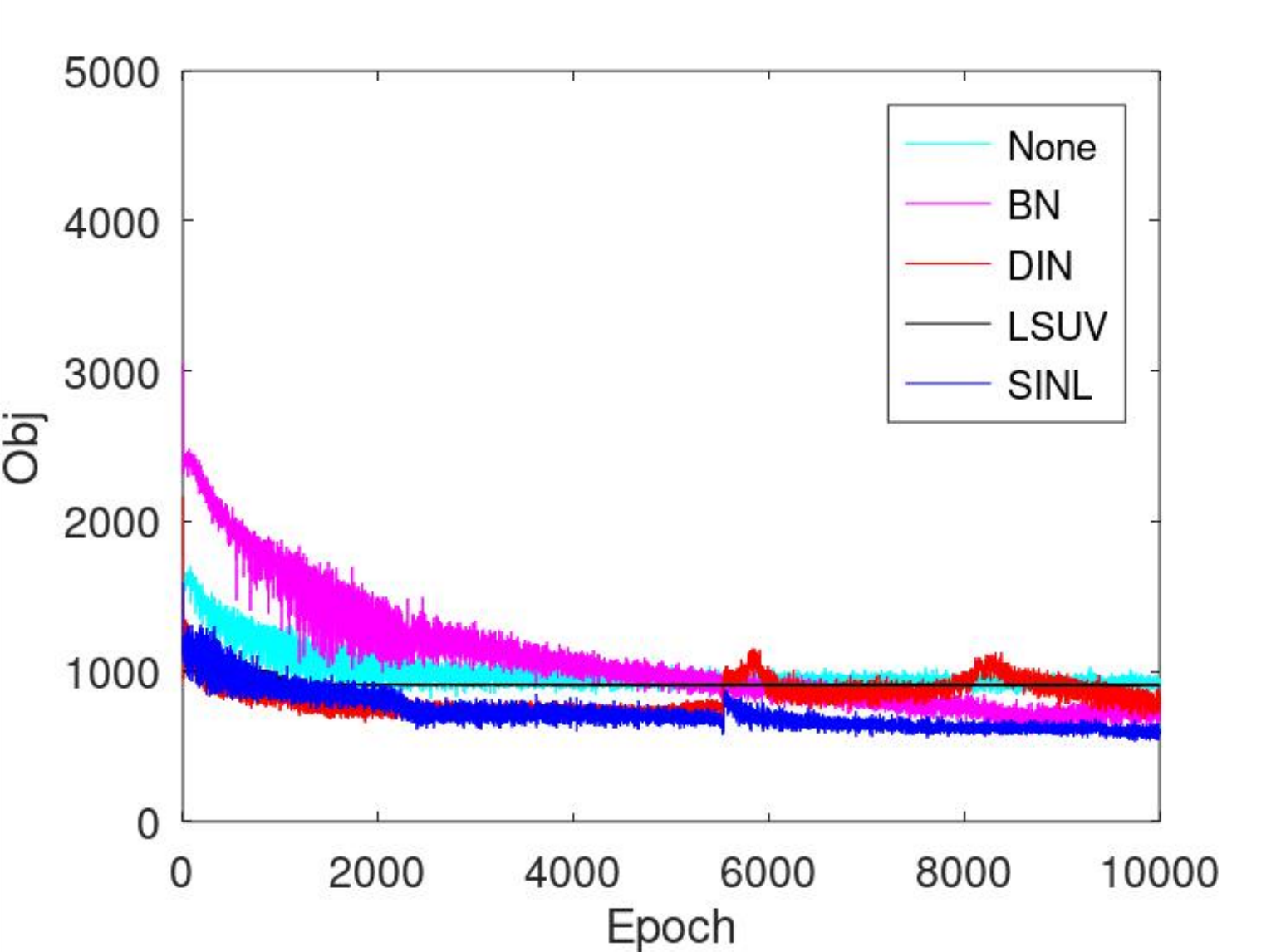}}
\caption{ The obtained objectives associated with the iterative epochs on the different data sets. (a) Coil 20 (b) Monkey (c) Letter. }
\end{figure*}
% &&&&&&&&&&&&&&&&&&&&&&&&&&&&&&&&&&&&&&&&&&&&&&&&&
\section{Experiments}
The proposed shrinkage initialization method (SINL) is evaluated and tested in this section, while several state-of-the-art algorithms are involved, including batch normalization (BN) \cite{Ioffe15BN}, dynamic initialization of nonlinear learning \cite{Saxe14DIN}, the layer-sequential unit-variance initialization \cite{Mishkin16LSUV}. Furthermore, the neural learning with none of initialization is adopted as the base line algorithm. To observe the natural influence of initialization and make the neural learning be smooth, the Sigmoid activation function is adopted in all layers of network. Several artificial data sets are referred as the experimental platform for deep learning, including Coil 20 \cite{Nene96Coil}, Monkey \cite{Mario10Monkey}, and Letter \cite{Dua17Letter}. For convenience and efficiency, three layers are referred in all neural learning methods, and the transform dimensionality of median layers are set to be 10 and 500 sequentially.

For neural learning methods associated with different initialization approaches, 10,000 epochs are performed by following the standard neural learning approach. Then, the obtained accuracy and objectives of each epoch are recorded as the temporal results, as well as the updated transformation of each layer of networks. 
%The obtained transformation of different neural learning methods is given in Fig. 2. As illustrated, the obtained transformation from the proposed SINL method is quite different from other ones. Furthermore, the similar transformation is achieved by the methods associated with the similar approaches. 
The obtained transformation of shrinkage initialization is given in Fig. 2. As illustrated, the obtained transformation of different epochs are quite similar with each other, which implies that neural learning is under the influence of initialization indeed. 
With respect to the obtained results, the random initialization is quite similar to the BN method, which produces normalized variances based on the random start of neurons, while DIN also hold the similar transformation with the orthogonal assignments. With the normalization stage, the results obtained by LSUV presents the sparse characteristics of transformation. Note that, the proposed SINL presents the combined patterns of sequential orthogonal rotation and normalization initialization.

In addition, the obtained accuracy and objectives are given in Fig. 3 and Fig. 4 respectively. According to the obtained results, each neural learning approach is able to achieve the upgrading accuracy as the increasing epochs, and no outstanding method still holds the superior results during experiments. Furthermore, DIN presents the robust performance on the Monkey data set, while BN approach is more ideal than other methods on the Coil 20 data set. In other words, both the initialization and the platform contribute influences to the performance of neural learning either. Note that, the proposed method is still able to achieve the stable performance compared with other methods and can obtain the comparable results to the best. In terms of the optimization of networks, the decline tendency is able to be achieved by all methods in a few epochs, and stable decreasing results can be obtained. The best optimized results are obatined by SINL on the Coil 20, while the nearly best performance is obtained on the Monkey. Furthermore, it is shown that, the performance of BN is unstable with respect to different data sets, though simple normalization is adopted. The similar situation also occurs for the DIN approach, which adopt the normalization as the refined stage of initialization derived from the idea of BN. 
%%%%%%%%%%%%%%%%%%%%%%%%%%%%%%%%%%%%%%%%%%
For the letter data set, the similar results are obtained compared with the results from Coil 20 data set. More specifically, the DIN and SINL present the decline tendency faster than other methods and quite close accuracy are obtained during training of neural network, while BN can obtain the optimistic results gradually with increasing epochs. Note that, the BN initialization based neural learning reaches the decline slowly compared with other methods, due to the low dimensionality of letter data set. Particularly, both DIN and SINL initialization based neural learning is able to give outstanding performance for targets of approximation.

\section{Conclusion}
The advances of digital life have largely made the broad applications of intelligent systems, which mainly relied on the adaptive handling of information. As a general solution, neural learning is able to train a complicated network with respect to the specific targets, while backpropagation is adopted in each layer of networks. Furthermore, it is known that the performance of networks is promised to be attained with the specifically defined initialization, the structures of neuron layers and the activation functions. Nevertheless, the initialization of networks normally suffers from the overfitting and nonuniform distribution, and batch normalization has been the most outstanding solution. 
%%%%%%%%%%%%%%%%%%%%%%%%%%%%%%%%%%%%%%%%%%
In this work, an improved approach to initialization of fully connected networks is presented, which is adaptable for generalized initialization of networks with random neurons. 
%Rather than the existing solutions, the proposed method optimizes the original transformation of each layer in a shrinkage sequence of boundary of network, and orthogonal rotation of the bridge connection is inferred and adopted to update of neurons. 
As a consequence, the proposed method is able to be competent for generalized initialization of networks, while light complexity is required. Furthermore, smooth neural learning is adopted in this work to disclose the natural influence of initialization, and diverse impacts are to be avoided for observation. Experimental results on several data sets demonstrate that, the proposed initialization method is able to achieve robust performance compared with other approaches, while the obvious diversity of optimized transformation can be reserved.

%\section{Acknowledgments}
\begin{acks}
This work was partly supported by Innovation and Talent Foundation of Guangxi Province of China (RZ1900007485, AD19110154). 
%The author Miao Cheng received the funding. 
The funder had no role in study design, data collection and analysis, decision to publish, or preparation of the manuscript. The corresponding author of this work is Miao Cheng.
\end{acks}

% %%%%%%%%%%%%%%%%%%%%%%%%%%%%%%%%%%%%%%%%%%%%%%%%%


\begin{thebibliography}{8}

\bibitem{Zhou14BigData}
Z. Zhou, N. Chawala, Y. Jin, G. Williams. 2014. Big data opportunities and challenges: Discussion from Data Analytics Perspectives. IEEE Computational Intelligence Magazine 9, 4 (Nov 2014), 62-74.

\bibitem{Bengio13RL}
Y. Bengio, A. Courville, and P. Vincent. 2013. Representation Learning: A Review and New Perspective. IEEE Trans. Pattern Analysis and Machine Intelligence 35, 8 (Aug 2013), 1798-1828.

\bibitem{Hinton06NN}
G. Hinton and R. Salakhutdinov. 2016. Reducing the Dimensionality of Data with Neural Networks. Science 313, 5786 (July 2006), 504-507.

\bibitem{Hastie11ESL}
T. Hastie, R. Tibshirani, and J. Friedman. 2016. The Elements of Statistical Learning: Data Mining, Inference, and Predication (2nd Ed.). Springer, Stanford, USA.

\bibitem{Bishop23DLFC}
C. M. Bishop, H. Bishop. 2023. Deep Learning: Foundations and Concepts (1st Ed.), Springer Press.

\bibitem{Goodfellow2016DL}
I. Goodfellow, Y. Bengio, A. Courville. 2016. Deep Learning. MIT Press.

\bibitem{Glorot10DFNN}
X. Glorot and Y. Bengio. 2010. Understanding the Difficulty of Training Deep Feedforward Neural Networks. In Proceedings of the 13th International Conference on Artificial Intelligence and Statistics, Sardinia, Italy.

\bibitem{Haykin16NNLM}
S. Haykin. 2016. Neural Networks And Learning Machines (3rd Ed.). Pearson Education.

\bibitem{Aggarwal23NNDL}
C. C. Aggarwal. 2023. Neural Networks and Deep Learning: A Textbook (2nd Ed.). Springer Press.

\bibitem{Taylor17MNN}
M. Taylor, M. Koning. 2017. The Math of Neural Networks. Blue Windmill Media.

\bibitem{Taylor17NNVIB}
M. Taylor, M. Koning. 2017. Neural Network: A Visual Introduction For Beginners. Blue Windmill Media.

\bibitem{Cuda00PC}
R. O. Cuda, P. E. Hart. 2000. Pattern Classification (2nd Ed.). Wiley.

\bibitem{Chan15PCANet}
T. Chan, K. Jia, S. Gao, J. Lu, Z. Zeng, and Y. Ma. 2015. PCANet: A Simple Deep Learning Baseline for Image Classification? IEEE Trans. Image Processing 24, 12 (Dec 2015) 5017-5032.

\bibitem{Krizhevsky12CNN}
A. Krizhevsky, I. Sutskever, G. Hinton. 2017. Imagenet Classification with Deep Convolutional Neural Networks. Communications of the ACM 60, 6 (May 2017), 84-90. 

\bibitem{Sercu16DCNN}
T. Sercu, C. Puhrsch, B. Kingsbury, Y. LeCun. 2016. Very Deep Multilingual Convolutional Neural Networks for LVCSR. In Proceedings of the 41st International Conference on Acoustics, Speech and Signal Processing. Shanghai, China.

\bibitem{Bengio94RNN}
Y. Bengio, P. Simard, and P. Frasconi. 1994. Learning Long-term Dependencies with Gradient Descent is Difficult. IEEE Trans. Neural Networks 5, 2 (March 1994), 157-166.

\bibitem{Bishop11PRML}
C. M. Bishop. 2011. Pattern Recognition and Machine Learning. Springer.

%\bibitem{Sutskever11RNN}
%I. Sutskever, J. Martens, and G. E. Hinton. 2011. Generating Text with Recurrent Neural Networks. In: Proc. International Conference on Machine Learning, Washington, USA.

%\bibitem{Hochreiter97LSTM}
%S. Hochreiter, J. Schmidhuber. 1997. Long Short-Term Memory. Neural Computation 9, 8 (Nov 1997), 1735-1780.

\bibitem{Hinton15DKNN}
G. Hinton, O. Vinyals, J. Dean. 2015. Distilling the Knowledge in a Neural Network. In Proceedings of the 28th International Conference on Advances in Neural Information Processing Systems: Deep Learning and Representation Learning Workshop, Montreal, Canada.

\bibitem{Bengio07GT}
Y. Bengio, P. Lamblin, D. Popovici, and H. Larochelle. 2006. Greedy Layer-wise Training of Deep Networks. In Proceedings of the 19th International Conference on Advances in Neural Information Processing Systems. Vancouver, Canada.

\bibitem{Srivastava14Dropout}
N. Srivastava, G. Hinton, A. Krizhevsky, I. Sutskever, R. Salakhutdinov. 2014. Dropout: A Simple Way to Prevent Neural Networks from Overfitting. The Journal of Machine Learning Research 15, 1 (Jan 2014), 1929-1958.

\bibitem{Ioffe15BN}
S. Ioffe, C. Szegedy. 2015. Batch Normalization: Accelerating Deep Network Training by Reducing Internal Covariate Shift. In Proceedings of the 32nd International Conference on Machine Learning. Lille, France.

\bibitem{Saxe14DIN}
A. Saxe, J. L. McClelland, S. Ganguli. 2014. Exact Solutions to the Nonlinear Dynamics of Learning in Deep Linear Neural Networks. In Proceedings of the 2nd International Conference on Learning Representation. Banff, Canada.

\bibitem{Mishkin16LSUV}
D. Mishkin, J. Matas. 2016. All You Need Is A Good Init. In Proceedings of the 4th International Conference on Learning Representation. San Juan, Puerto Rico.


\bibitem{Graham14FMP}
B. Graham. 2014. Fractional Max-Pooling. https://arxiv.org/abs/1412.6071.

\bibitem{Murray14GMP}
N. Murray, F. Perronnin. 2014. Generalized Max Pooling. In Proceedings of International Conference on Computer Vision and Pattern Recognition, Columbus, USA.

\bibitem{Goodfellow13Maxout}
I. Goodfellow, D. W. Farley, M. Mirza, A. Courville, Y. Bengio. 2013. Maxout networks. In Proceedings of the 30th International Conference on Machine Learning. Atlanta, USA.

\bibitem{Clevert16ELU}
D. A. Clevert, T. Unterthiner, S. Hochreiter. 2016. Fast and Accurate Deep Network Learning by Exponential Linear Units (ELUs). In Proceedings of the 4th International Conference on Learning Representation. San Juan, Puerto Rico.

\bibitem{Chang15BN}
J. R. Chang, and Y. S. Chen. 2015. Batch-normalized Maxout Network in Network. https://arxiv.org/abs/1511.02583.

\bibitem{Cheng19SPA}
M. Cheng, W. Yang, Y. Li, S. Zhang, A. C. Tsoi, and Y. Y. Tang. 2019. Sequential Pattern Learning via Kernel Alignment. In Proceedings of the 11th International Conference on Advanced Computational Intelligence. Guilin, China.

\bibitem{Cheng22TANMF}
M. Cheng, F. Zhou, and J. Wu. 2021. Online Nonnegative Matrix Factorization with Temporal Affinity. In Proceedings of the 6th International Conference on Signal and Image Processing. Suzhou, China.

\bibitem{Cheng10LDSE}
M. Cheng, B. Fang, Y. Y. Tang, T. Zhang, and J. Wen. 2010. Incremental Embedding and Learning in the Local Discriminant Subspace With Application to Face Recognition. IEEE Trans. Systems, Man, and Cybernetics 40, 5 (Sep 2010), 580-591.

\bibitem{Liao16NL}
Z. Liao, G. Carneiro. 2016. On the Importance of Normalisation Layers in Deep Learning with Piecewise Linear Activation Units. In Proceedings of Winter Conference on Applications of Computer Vision, Lake Placid, USA.

\bibitem{Kuo22DSDNet}
P. H. Kuo, J. Pan, S. Y. Chien, M. H. Yang. 2022. Learning Discriminative Shrinkage Deep Networks for Image Deconvolution, In Proceedings of the 17th European Conference on Computer Vision, Tel Aviv, Israel.

\bibitem{Golub13MC}
G. H. Golub, C. F. V. Loan. 2013. Matrix Computations (4th Ed.). Johns Hopkins University Press.

\bibitem{Cheng11MMC}
M. Cheng, Y. Y. Tang. 2011. Nonparametric Feature Extraction via Direct Maximum Margin Alignment. In Proceedings of the 10th International Conference on Machine Learning and Applications. Honolulu, USA.

\bibitem{Cheng18MMC}
M. Cheng, Z. Liu, H. Zou, A. C. Tsoi. 2018. A Family of Maximum Margin Criterion for Adaptive Learning. In Proceedings of 25th International Conference on Neural Information Processing. Siem Reap, Cambodia.

\bibitem{Nene96Coil}
S. A. Nene, S. K. Nayar and H. Murase. 1996. Columbia Object Image Library (COIL-20), Technical Report CUCS-005-96.

\bibitem{Mario10Monkey}
S. Mario, R. Renard, G. Montoya, J. Zhang, S. Loaiciga, The 10 Monkey Species, https://www.kaggle.com/slothkong/10-monkey-species.

\bibitem{Dua17Letter}
D. Dua, C. Graff. 2017. UCI Machine Learning Repository: Letter Recognition Data, https://archive.ics.uci.edu/dataset/59/letter+recognition.



\end{thebibliography}
\end{document}